\pdfoutput=1

\documentclass[11pt]{article}

\usepackage{EMNLP2022}
\usepackage{times}
\usepackage{latexsym}

\usepackage[T1]{fontenc}

\usepackage[utf8]{inputenc}

\usepackage{microtype}

\usepackage{inconsolata}

\usepackage{microtype}
\usepackage{amsmath}
\usepackage{amssymb}
\usepackage{multirow}
\usepackage{booktabs}

\usepackage{graphicx}
\usepackage{epstopdf}

\usepackage{amsmath}
\usepackage{CJKutf8}
\usepackage{url}

\title{WeTS: A Benchmark for Translation Suggestion}

\author{Zhen Yang,  Fandong Meng, Yingxue Zhang, Ernan Li, and Jie Zhou \\
   Pattern Recognition Center, WeChat AI, Tencent Inc, Beijing, China \\
  {\tt \{zieenyang, fandongmeng, yxuezhang, cardli, withtomzhou\}@tencent.com}}
  
\begin{document}
\begin{CJK}{UTF8}{gbsn} 
\maketitle
\begin{abstract}
Translation suggestion (TS), which provides alternatives for specific words or phrases given the entire documents generated by machine translation (MT), has been proven to play a significant role in post-editing (PE). There are two main pitfalls for existing researches in this line. First, most conventional works only focus on the overall performance of PE but ignore the exact performance of TS, which makes the progress of PE sluggish and less explainable; Second, as no publicly available golden dataset exists to support in-depth research for TS, almost all of the previous works conduct experiments on their in-house datasets or the noisy datasets built automatically, which makes their experiments hard to be reproduced and compared. To break these limitations mentioned above and spur the research in TS,  we create a benchmark dataset, called \emph{WeTS}, which is a golden corpus annotated by expert translators on four translation directions. Apart from the golden corpus, we also propose several methods to generate synthetic corpora which can be used to improve the performance substantially through pre-training. As for the model, we propose the segment-aware self-attention based Transformer for TS. Experimental results show that our approach achieves the best results on all four directions, including English-to-German, German-to-English, Chinese-to-English, and English-to-Chinese. Codes and corpus can be found at \url{https://github.com/ZhenYangIACAS/WeTS.git}.
\end{abstract}

\section{Introduction}
Computer-aided translation (CAT) \cite{barrachina2009statistical,green2014human,knowles2016neural,santy2019inmt} has attained more and more attention for its promising ability in combining the high efficiency of machine translation (MT) \cite{cho2014learning,bahdanau:14,vaswani2017attention} and high accuracy of human translation (HT). A typical way for CAT tools to combine MT and HT is PE \cite{green2013efficacy,zouhar2021neural}, where the human translators are asked to provide alternatives for the incorrect word spans in the results generated by MT. To further reduce the post-editing time, researchers propose to apply TS into PE, where TS provides the sub-segment suggestions for the annotated incorrect word spans in the results of MT, and their extensive experiments show that TS can substantially reduce translators' cognitive loads and the post-editing time \cite{wang2020touch,lee2021intellicat}.

As there is no explicit and formal definition for TS, we observe that some previous works similar or related to TS have been proposed \cite{alabau2014casmacat,santy2019inmt,wang2020touch,lee2021intellicat}. However, there are two main pitfalls for these works in this line. First, most conventional works only focus on the overall performance of PE but ignore the exact performance of TS. This is mainly because the golden corpus for TS is relatively hard to collect. As TS is an important sub-module in PE, paying more attention to the exact performance of TS can boost the performance and interpretability of PE. Second, almost all of the previous works conduct experiments on their in-house datasets or the noisy datasets built automatically, which makes their experiments hard to be followed and compared. Additionally, experimental results on the noisy datasets may not truly reflect the model's ability on generating the right predictions, making the research deviate from the correct direction. Therefore, the community is in dire need of a benchmark for TS to enhance the research in this area.

To address the limitations mentioned above and spur the research in TS, we make our efforts to construct a high-quality benchmark dataset with human annotation, named \emph{WeTS},\footnote{\emph{WeTS}: We Establish a benchmark for Translation Suggestion} which covers four different translation directions. As collecting the golden dataset is expensive and labor-consuming, we further propose several methods to automatically construct synthetic corpora, which can be utilized to improve the TS performance through pre-training. As for the model, we for the first time propose the segment-aware self-attention based Transformer for TS, named SA-Transformer, which achieves superior performance to the naive Transformer \cite{vaswani2017attention}. The main contributions of this paper are summarized as follows:
\begin{itemize}
    \item We construct and share a benchmark dataset for TS on four translation directions.
    \item We provide strong baseline models for this community. Specifically, we make a detailed comparison between the Transformer-based and XLM-based models, and propose the segment-aware self-attention based Transformer for TS, which achieves the best results on the benchmark dataset.
    \item We thoroughly investigate different ways for building the synthetic corpora. Since constructing the golden corpus is expensive and labor-consuming, it is very essential and promising to build the synthetic corpora by making full use of the parallel corpus of MT.  
    \item We conduct extensive experiments and provide deep analyses about the strengths and weaknesses of the proposed approach, which are expected to give some insights for further researches on TS.
\end{itemize}

\section{WeTS}
This section introduces the proposed dataset \emph{WeTS}. To make the constructing process understood easily, we first formally define the task of TS.
\subsection{Translation Suggestion}
Given the source sentence $\boldsymbol{x}=(x_1, \ldots, x_s)$, the translation sentence $\boldsymbol{m}=(m_1,\ldots,m_t)$, the incorrect words or phrases $\boldsymbol{w}=\boldsymbol{m}_{i:j}$ where $1 \le i \le j \le t$, and the correct alternative $\boldsymbol{y}$ for $\boldsymbol{w}$, the task of TS is optimized to maximize the conditional probability of $\boldsymbol{y}$ as follows:
\begin{equation}
    \label{eq: definition}
    P(\boldsymbol{y}|\boldsymbol{x}, \boldsymbol{m}^{-\boldsymbol{w}}, \theta)
\end{equation}
where $\theta$ represents the model parameter, and $\boldsymbol{m}^{-\boldsymbol{w}}$ is the masked translation where the incorrect word span $\boldsymbol{w}$ is replaced with a placeholder. \footnote{$\boldsymbol{w}$  is null if $i$ equals $j$, and the model will predict whether some words need to be inserted in position $i$.}

\begin{table}[ht]
			\centering
			\resizebox{0.95 \columnwidth}{!}{
				\begin{tabular}{c|ccc}
					\toprule[2pt]
					 Translation Direction & Train & Valid & Test \\
					\midrule[1pt]
			        En$\Rightarrow$De & 14,957 & 1000 & 1000 \\
			        De$\Rightarrow$En & 11,777& 1000 & 1000 \\
			        Zh$\Rightarrow$En & 21,213& 1000 & 1000  \\
			        En$\Rightarrow$Zh & 15,769& 1000 & 1000  \\
					\bottomrule[2pt]
				\end{tabular}}
					\caption{\label{tab:statics of WeTs} The sizes for cases in train/valid/test sets. ``En$\Rightarrow$De” refers to the direction of English-to-German, and ``En$\Rightarrow$Zh” refers to English-to-Chinese.}
\end{table}
\begin{figure*}[t]
    \centering
    \includegraphics[scale=0.85]{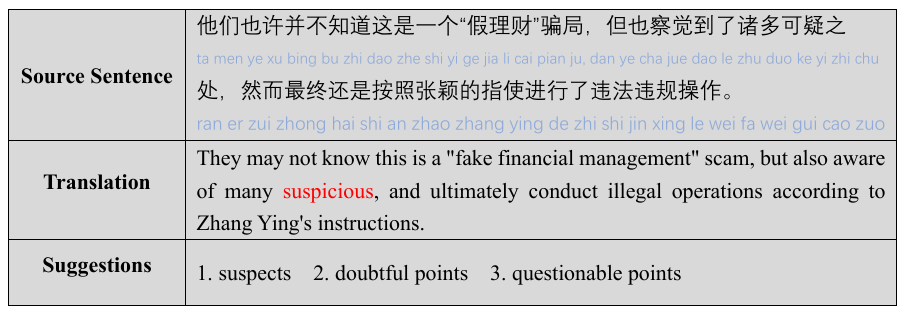}
    \caption {One training example in \emph{WeTS}. For the incorrect word "suspicious" (in red color), there are three correct suggestions. For readability, we also provide the Chinese pinyin format for the Chinese sentence (in blue color). }
    \label{fig:example}
\end{figure*}
\subsection{Dataset}
This sub-section describes the annotation guidelines and construction process for \emph{WeTS}, which is a golden corpus for four translation directions, including English-to-German, German-to-English, Chinese-to-English and English-to-Chinese.
\paragraph{Annotation Guidelines}
It is non-trivial for annotators to locate the incorrect word spans in the MT sentence. The main difficulty is that, the concept of ``translation error" is ambiguous and each translator has his own understanding about translation errors. To easier the annotation workload and reduce the possibility of making errors, we group the translation errors on which we aim to focus into three macro categories:
\begin{itemize}
    \item Under-translation or over-translation: While the problem of under-translation or over-translation has been alleviated with the popularity of Transformer, it is still one of the main mistakes in NMT and seriously destroys the readability of the translation.
    \item Semantic errors:  For the semantic error, we mean that some source words are incorrectly translated according to the semantic context, such as the incorrect translations for entities, proper nouns, and ambiguous words. Another branch of semantic mistake is that the source words or phrases are only translated superficially and the semantics behind are not translated well.
    \item Grammatical or syntactic errors: Such errors usually appear in translations of long sentences, including the improper use of tenses, passive voice, syntactic structures, etc.
\end{itemize}
 Another key rule for translators is that annotating the incorrect span as local as possible, as generating correct alternatives for long sequences is much harder than that of shorter sequences. 
 
\paragraph{Data Construction}
As the starting point, we collect the monolingual corpora for English and German from the raw Wikipedia dumps, and extract Chinese monolingual corpus from various online news publications. We first clean the monolingual corpora with a language detector to remove sentences belonging to other languages.\footnote{\url{https://github.com/Mimino666/langdetect}} For all monolingual corpora, we remove sentences that are shorter than 20 words or longer than 80 words. In addition, sentences which exist in the available parallel corpora are also removed. Then, we get the translations by feeding the cleaned monolingual corpus into the corresponding fully-trained NMT model.  The NMT models for English-German language pairs are trained on the parallel corpus of WMT14 English-German. For Chinese-English directions, the NMT models are trained with the combination between the WMT19 English-Chinese\footnote{\url{https://www.statmt.org/wmt19/translation-task.html}} and the same amount of in-house corpus. \footnote{We have released the models and inference scripts utilized here to make our results easy reproduced.}

Finally, the translators are required to mark the incorrect word spans in the translation sentence and 
provide at least one alternative for each incorrect span, by using the annotation guidelines. The team is composed by eight annotators with high expertise in translation and each example has been assigned to three experts. There are two phases of agreement computations. In the first phase,
an annotation is considered in agreement among the experts if and only if they capture the same incorrect word spans. If one annotation passes the first agreement computation, it will be assigned to other three experts in charge of selecting the right alternatives from the previous annotation. In the second phase of agreement computation, an annotation is considered in agreement among the experts if and only if they select the same right alternatives. With the two-phase agreement checking, we ensure the high quality of the annotated examples. For the annotated examples with multiple incorrect word spans, we can extract multiple examples which have the same source and translation sentences, but different incorrect word span and the corresponding suggestions. Finally the extracted examples are randomly shuffled and then split into the training, validation and test sets.\footnote{To keep the fairness of \emph{WeTS}, we ensure the examples among the training, validation and test sets have different source and translation sentences.} One training example for the translation direction of Chinese-to-English is presented in Figure \ref{fig:example} and the sizes for the train/valid/test sets in \emph{WeTS} are collected in Table \ref{tab:statics of WeTs}. Readers can find the detailed statistics about \emph{WeTS} in Appendix \ref{app:statistics}.


\begin{figure*}[htb]
    \centering
    \includegraphics[scale=0.49]{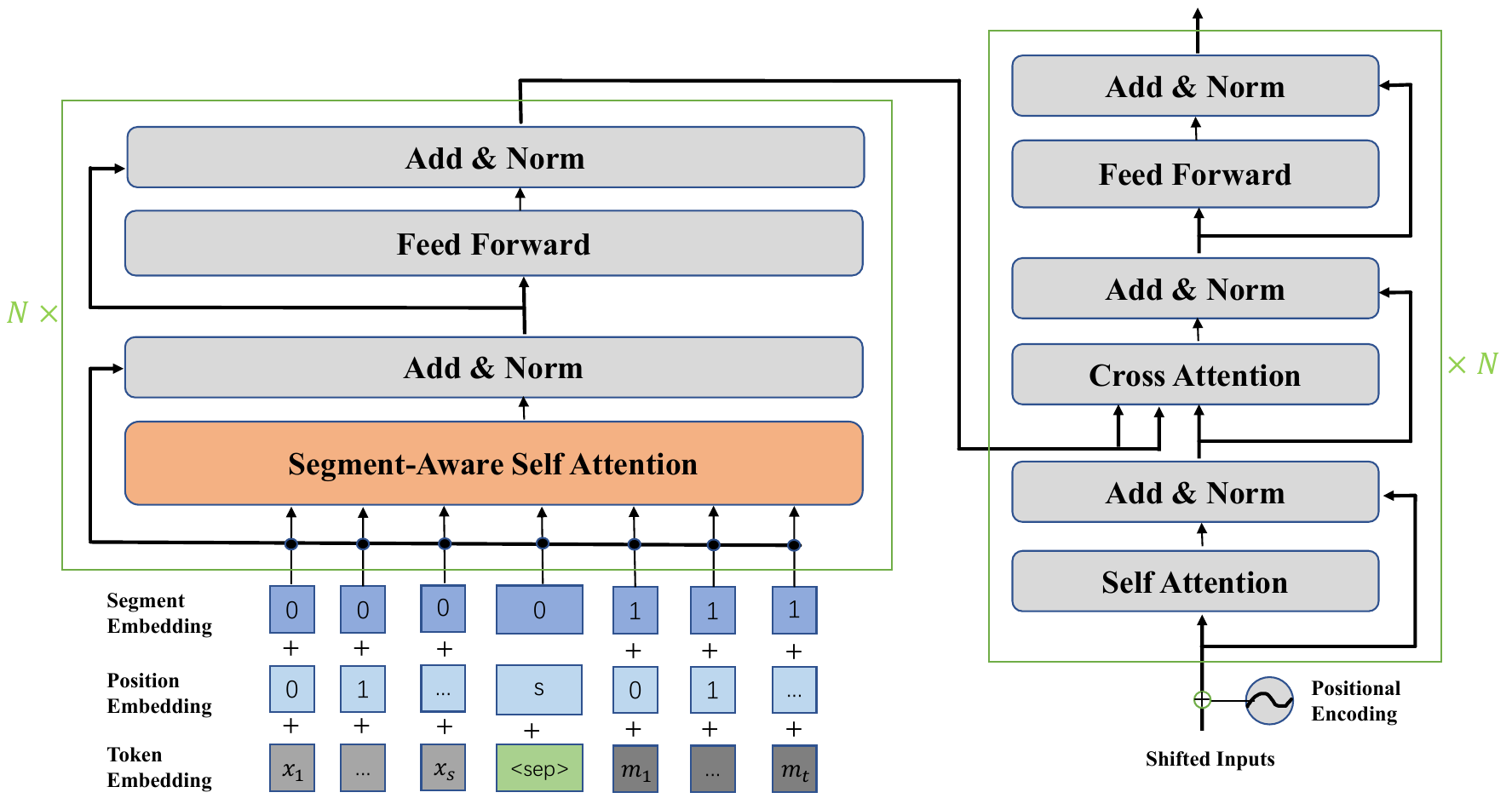}
    \caption {The whole architecture of the proposed  SA-Transformer. The $s$ and $t$ are the lengths for $\boldsymbol{x}$ and $\boldsymbol{m}^{-\boldsymbol{w}}$ respectively. $N$ denotes the layers for the encoder and decoder.}
    \label{fig:model}
\end{figure*}

\section{Construct Synthetic Corpus}
Since constructing the golden corpus is expensive and labor-consuming, automatically building the synthetic corpus is very promising. In this section, we describe several ways for constructing synthetic corpus for TS based on the parallel corpus of MT.
\subsection{Sampling on Golden Parallel Corpus}
\label{sec:sampling on golden}
Sampling on the golden parallel corpus of MT is the most straightforward and simplest way for constructing synthetic corpus for TS. Given the sentence pair $(\boldsymbol{x}, \boldsymbol{r})$ in the parallel corpus of MT, where $\boldsymbol{x}$ is the source sentence and $\boldsymbol{r}$ is the corresponding target sentence, we denote $\boldsymbol{r}^{\backslash i:j}$ as a masked version of $\boldsymbol{r}$ where its fragment from position $i$ to $j$ is replaced with a placeholder ($1 \le i \le j \le |\boldsymbol{r}|$). The $\boldsymbol{r}^{i:j}$ denotes the fragment of $\boldsymbol{r}$ from position $i$ to $j$. We treat $\boldsymbol{r}^{i:j}$ and $\boldsymbol{r}^{\backslash i:j}$ as the correct alternative ($\boldsymbol{y}$ in Equation \ref{eq: definition}) and masked translation ($\boldsymbol{m}^{-\boldsymbol{w}}$ in Equation \ref{eq: definition}) respectively. In this approach, the masked translation in each example is part of the golden 
target sentence. However, in production, the TS model needs to predict the correct suggestions based on the context of the machine translated sentence. Therefore, the mismatch of distribution between the golden target sentence and machine translated sentence is the potential pitfall for this approach.

\subsection{Sampling on Pseudo Parallel Corpus}
The second approach we apply to construct the synthetic corpus for TS is sampling on the pseudo parallel corpus of MT. Given the source sentence $\boldsymbol{x}$ and the MT model $\boldsymbol{T}_{\theta}$, we first get the translated sentence $\boldsymbol{\tilde y}$ by feeding $\boldsymbol{x}$ into $\boldsymbol{T}_{\theta}$, and $(\boldsymbol{x}, \boldsymbol{\tilde y})$ is treated as the pseudo sentence pair. Then, we perform sampling on $(\boldsymbol{x},\boldsymbol{\tilde y})$ as what we do on $(\boldsymbol{x},\boldsymbol{r})$ in Section \ref{sec:sampling on golden}. Compared to the approach of sampling on the golden parallel corpus, sampling on the pseudo parallel corpus can address the problem of distribution mismatch mentioned in Section  \ref{sec:sampling on golden} and it works without relying on the golden parallel corpus. However, the suggested alternatives may be in poor quality since they are parts of the translated sentences. 

\subsection{Extracting with Word Alignment}
Considering the shortcomings of the two previous approaches, we investigate the third approach where we conduct the word alignment between the machine translation and the golden target sentence, and then extract the synthetic corpus for TS based on the alignment information. Given the sentence triple $(\boldsymbol{x}, \boldsymbol{\tilde y}, \boldsymbol{r})$, we perform word alignment between $\boldsymbol{\tilde y}$ and $\boldsymbol{r}$, and extract the aligned phrase table.\footnote{We test two different ways for performing word alignment, including fast-align \cite{dyer2013simple} and TER \cite{snover2006study}, and the experimental results show that fast-align performs better.} For phrase $\boldsymbol{\tilde y}^{i:j}$ in $\boldsymbol{\tilde y}$ and its aligned phrase $\boldsymbol{r}^{a:b}$ in $\boldsymbol{r}$, we denote $\boldsymbol{\tilde y}^{\backslash i:j}_{r^{a:b}}$ as the modified version of $\boldsymbol{\tilde y}$ where the phrase $\boldsymbol{\tilde y}^{i:j}$ is replaced with $\boldsymbol{r}^{a:b}$. If $\boldsymbol{r}^{a:b}$ is not identical to $\boldsymbol{\tilde y}^{i:j}$ and the perplexity of $\boldsymbol{\tilde y}^{\backslash i:j}_{r^{a:b}}$ is lower than that of $\boldsymbol{\tilde y}$ with a margin no less than $\beta$, we treat $\boldsymbol{\tilde y^{\backslash i:j}}$ and $\boldsymbol{r}^{a:b}$ as the masked translation and the correct alternative respectively.\footnote{To measure the sentence perplexity, we use kenlm (\url{https://github.com/kpu/kenlm}), the widely used toolkit for n-gram language model.} $\beta$ is a hyper-parameter to control the threshold of the margin. While this approach has achieved much improvement compared to the previous approaches, we still notice that the errors in the extracted alignment information may introduce some noises. Recent advances in neural word aligners may provide more accurate alignments \cite{sabet2020simalign,Lai2022cross}, and we leave this in our future work.

\section{The Model}
In this section, we describe the proposed model, i.e., SA-Transformer, and the whole architecture is illustrated as Figure \ref{fig:model}.
\subsection{Inputs}
Given the source sentence $\boldsymbol{x}$ and the masked translation $\boldsymbol{m}^{-\boldsymbol{w}}$, the input to the model is formatted as:
\begin{equation}
\left [\boldsymbol{x};\langle sep \rangle; \boldsymbol{m}^{-\boldsymbol{w}} \right]
\end{equation}
where $\left[\cdot;\cdot\right]$ means concatenation, and $\langle sep \rangle$ is a special token used as a delimiter.
The position for each segment in the input is calculated independently and we use the segment embedding to distinguish each segment from others. The representation for each token in the input is the sum of its token embedding, position embedding and segment embedding. 

\subsection{Segment-aware Self-attention}
The naive Transformer applies the self-attention to extract the higher-level information from the token representations in the lower layer without distinguishing tokens in each segment from those in other segments explicitly. The attention matrix in the self-attention is typically calculated as:
\begin{equation}
\frac{\boldsymbol{Q}\boldsymbol{W^{Q}}(\boldsymbol{K}\boldsymbol{W}^K)}{\sqrt{d_x}}
\end{equation}
where the $\boldsymbol{Q}$ and $\boldsymbol{K} \in R^{s*d_x}$ are identical in the encoder, $\boldsymbol{W^{Q}}$ and $\boldsymbol{W^{K}} \in R^{d_x * d_x}$ are the projection matrix, $d_x$ is the dimension of the word embedding.
However, the inputs for TS contain tokens from different segments, i.e., the source sentence, masked translation, and the hints if provided, and the tokens in each segment are expected to be distinguished from those in other segments since they provide different information for the model's prediction. 
While the segment embedding in token representations has played the role for distinguishing tokens from different segments, its information has been mixed with the word embedding and diluted with the information flow. 
With this consideration, we propose the segment-aware self-attention by further injecting the segment information into the self-attention to make it perform differently according to the segment information of the tokens. Formally, the attention matrix in the proposed segment-aware self-attention is calculated as:
\begin{equation}
\frac{(\boldsymbol{E}_{seg} \cdot \boldsymbol{Q}) \boldsymbol{W^{Q}}((\boldsymbol{E}_{seg} \cdot \boldsymbol{K})\boldsymbol{W}^K)}{\sqrt{d_x}}
\end{equation}
where $\boldsymbol{E}_{seg} \in \boldsymbol{R}^{s * d_x}$ is the segment embedding and $\cdot$ represents dot production.\footnote{We also tried to sum the $\boldsymbol{E}_{seg}$ with $\boldsymbol{Q}$ or $\boldsymbol{K}$, but we did not get any improvement.}

\subsection{Two-phase Pre-training}
We apply the pretraining-finetuning paradigm for training the proposed model. The pre-training process can be divided into two phases: In the first phase, we follow \citet{lee2021intellicat} to pre-train a XLM-R model with a modified translation language model objective on the monolingual corpus, and then utilize the pre-trained parameters of XLM-R to initialize the encoder of the proposed model.\footnote{For details about the first-phase pre-training, we refer the readers to the work of \citet{lee2021intellicat}.} In the second phase, we apply the combination of all the constructed synthetic corpus to pre-train the whole model. After pre-training, we finetune the model on the golden training set of \emph{WeTS}.

\begin{table*}[htb]
\centering
\resizebox{2.0 \columnwidth}{!}{
\begin{tabular}{@{}l|l|cccc@{}|cccc@{}}
\toprule
  \multirow{2}{*}{\#}   &     \multirow{2}{*}{Systems}     & \multicolumn{4}{c|}{BLEU}  & \multicolumn{4}{c}{BLEURT}      \\ 
    &       &Zh$\Rightarrow$En  & En$\Rightarrow$Zh  &De$\Rightarrow$En & En$\Rightarrow$De &Zh$\Rightarrow$En  & En$\Rightarrow$Zh  &De$\Rightarrow$En & En$\Rightarrow$De  \\\hline\hline
1 & XLM-R  &  21.25 & 32.48   & 27.40  & 25.12  & 40.17 & 52.05 & 37.21 & 36.40 \\
2 & Naive Transformer & 24.20  &  35.01 & 30.08  &  28.15 & 43.32 & 54.60 & 41.05 & 41.21 \\
3 & Dual-source Transformer  & 24.29  & 35.10  & 30.23 & 28.09 & 43.12 & 54.75 & 42.01 & 40.95\\
\hline
4 & \textbf{SA-Transformer (ours)} & \textbf{25.51*}   & \textbf{36.28*}  & \textbf{31.20*}   &    \textbf{29.48*} & \textbf{44.67*}& \textbf{56.48*} &\textbf{42.66} & \textbf{42.13*}      \\ \bottomrule
\end{tabular}}
 \caption{\label{tab:all} The main results on the four language pairs. The numbers with '*' indicate the significant improvement over the baseline of naive Transformer with $p<0.01$ under t-test.}
\end{table*}


\section{Experiments and Results}
We first describe the experimental settings, including datasets, pre-processing, and hyper-parameters; Then we introduce the baseline systems and report the main experimental results.
\subsection{Datasets and Pre-processing}
To make our results reproducible, we construct the synthetic corpora from the publicly available datasets provided by the WMT2019 and WMT2014 shared translation tasks. We use the full training set of the WMT14 English-German, which contains 4.5M sentence pairs. For the WMT19 Chinese-English dataset, we remove sentences longer than 200 words and get 20M sentence pairs. The NMT models utilized for constructing synthetic corpus are identical to the ones used for constructing \emph{WeTS}. For each translation direction, the source and target corpus are jointly tokenized into sub-word units with BPE \cite{sennrich-etal-2016-neural}. The source and target vocabularies are extracted from the source and target tokenized synthetic corpus respectively. During fine-tuning, we pre-process the golden corpus with the same tokenizer utilized in pre-training. Sizes for the constructed synthetic corpora and the details about pre-processing can be found in Appendix \ref{app:processing}.
\subsection{Hyper-parameters and Evaluation}
We take the Transformer-base \cite{vaswani2017attention} as the backbone of our model, and we use beam search with a beam size of 4 for searching the results. The proposed model is implemented based on the open-source toolkit fairseq.\footnote{\url{https://github.com/pytorch/fairseq}}  For evaluation, we report the scores of BLEU \cite{papineni2002bleu} and BLEURT \cite{sellam2020bleurt} on the test sets of \emph{WeTS}. Both of the two scores are calculated between the top-1 generated suggestion against the golden suggestions in the reference. BLEU is the widely used metric for calculating the n-gram precision between candidates and references. For the direction of English-to-Chinese, we report the character-level BLEU. For the other three directions, we report the case-sensitive BLEU on the de-tokenized sentences. In this paper, we utilize the script of \emph{multi-bleu.pl} as the evaluation tool for BLEU. BLEURT is the recently proposed metric which returns a score that indicates to what extent the candidate is fluent and conveys the meaning of the reference. For calculating BLEURT score, we directly utilized the released toolkit \emph{bleurt}.\footnote{\url{https://github.com/google-research/bleurt}} We refer the readers to the appendix \ref{app:settings} for details about the experimental settings.

\subsection{Baselines}
\paragraph{XLM-R.} The first baseline system we consider is the work of \citet{lee2021intellicat} who propose the TS system based on XLM-R \cite{conneau-etal-2020-unsupervised}. Following \citet{lee2021intellicat}, we re-implement the XLM-based TS model based on the open-source toolkit of XLM \cite{lample2019cross} with slight modification.
\paragraph{Naive Transformer.} We take the naive Transformer \cite{vaswani2017attention} as the second baseline and directly apply the implementation of fairseq.
\paragraph{Dual-source Transformer.} We finally consider the dual-source Transformer \cite{junczys2018ms} which applies two shared encoders to encode the source sentence and masked translation respectively. We re-implement the model based on the fairseq toolkit.

All of the baseline systems mentioned above are trained in the same way as our system.
\subsection{Main Results}
Table \ref{tab:all} shows the main results of our experiments. We can find that, compared to all of the baseline systems, the proposed SA-Transformer achieves the best results on all of the four translation directions. Compared with the XLM-based approach (comparing systems 2-4 with system 1), the Transformer-based approach can achieve substantial gains on the final performance. While the dual-source Transformer has a more complex model structure, it only achieves comparable results with the naive Transformer. We conjecture the main reason is that the dual-source Transformer does not model the interaction between the source and translation, as the source and translated sentences are encoded with two separate encoders in the dual-source Transformer. Compared to the naive Transformer, the proposed model achieves the improvement up to +1.3 BLEU points and + 1.5 BLEURT points on Chinese-to-English translation direction.

\section{Analysis}
This section provides detailed analysis about the proposed approach, and performance on two translation directions are presented for most of the following experiments.
\subsection{Effects of Synthetic Parallel Corpus}
In this paper, we propose three different ways for constructing synthetic corpus for the second-phase pre-training. A natural question is that how each of the synthetic corpus affects the performance. We investigate this problem by studying the performance on the English-to-Chinese direction with different synthetic corpus. We report both intermediate and final performances of the model, where fine-tuning is removed and applied respectively. Results are presented in Table \ref{tab:analysis corpus}. As shown in Table \ref{tab:analysis corpus}, we can find that the model trained on the combination of all three kinds of synthetic corpus achieves the best performance. The synthetic corpus constructed with word alignment contributes the most to the final performance. 
\begin{table}[htb]
\resizebox{0.98 \columnwidth}{!}{
\begin{tabular}{l|cc@{}}
\toprule
 \multirow{2}{*}{Systems}  & \multicolumn{2}{c}{En$\Rightarrow$Zh} \\ 
 & \multicolumn{1}{c}{w/o finetuning} & \multicolumn{1}{c}{w/ finetuning}  \\ \hline \hline
SA-Transformer & \textbf{29.76} & \textbf{36.28}\\
w/o on golden corpus & 26.64& 34.27\\
w/o on pseudo corpus &26.04 & 34.01\\
w/o with word alignment & 21.26& 28.42\\
\bottomrule
\end{tabular}}
\caption{\label{tab:analysis corpus} Results on the effects of synthetic corpus.}
\end{table}

\subsection{Study the Training Procedure}
We adopt the pretraining-finetuning paradigm for the model training, where the two-phase pre-training enhances the model's ability in modeling the general inputs and the fine-tuning further enhances the performance on the golden test sets. In this section, we aim to investigate how the training procedure affects the final performance. Table \ref{tab:analysis training} shows the experimental results. As Table \ref{tab:analysis training} shows, the model achieves very low BLEU scores, i.e., 6.70 in English-to-Chinese and 5.87 in English-to-German, if pre-training is not applied. This is mainly because that the golden corpus of \emph{WeTS} is too scarce to train a well-performed TS model. In the two-phase pre-training, the second-phase pre-training plays a more important role for the final performance, with a decrease of almost 20 BLEU score on English-to-Chinese translation direction if removed. Fine-tuning on the golden corpus of \emph{WeTS} substantially enhances the performance, with an improvement of almost 8 BLEU score on the English-to-German translation direction. 
\begin{table}[htb]
\resizebox{0.94 \columnwidth}{!}{
\begin{tabular}{l|c|c}
\toprule
System & En$\Rightarrow$Zh & En$\Rightarrow$De \\ \hline \hline
SA-Transformer & \textbf{36.28} & \textbf{29.48} \\
w/o fine-tuning & 29.76 & 21.44 \\
w/o pre-training & 6.70 & 5.87 \\
w/o first-phase pre-training & 34.63& 28.37 \\
w/o second-phase pre-training &16.85 & 14.14\\
\bottomrule
\end{tabular}}
\caption{\label{tab:analysis training} Results on the effects of training strategies.}
\end{table}
\subsection{Ablation Study on Model Structure}
To understand the importance of the model components, we perform an ablation study by training multiple versions of the model with some components removed or degenerated into the corresponding ones in the naive Transformer. We mainly test three components, including the independent position encoding, segment embedding, and the segment-aware self-attention. Results are reported in Table \ref{tab:ablation}. We find that the best performance is obtained with the simultaneous use of all test components. The most critical component is the segment-aware self-attention, which enables the model to perform a different calculations of self-attention according to the type of the input tokens. When we remove the segment embedding, we get 0.46 BLEU points decline on the English-to-Chinese translation direction. And when the segment-aware self-attention is removed, the decline can be as large as 0.77 BLEU points. These results indicate that the proposed segment-aware self-attention can provide more useful segment information.
\begin{table}[htb]
\resizebox{0.98 \columnwidth}{!}{
\begin{tabular}{l|c|c}
\toprule
System & En$\Rightarrow$Zh & En$\Rightarrow$De \\ \hline \hline
SA-Transformer & \textbf{36.28} & \textbf{29.48} \\
w/o independent position encoding &36.01 & 29.35\\
w/o segment embedding &35.82 & 29.01\\
w/o segment-aware self-attention & 35.51& 28.74 \\
\bottomrule
\end{tabular}}
\caption{\label{tab:ablation} Ablation study: ``w/o segment embedding" means that the segment embedding is not added into the token representation, but still inserted in the segment-aware self-attention.}
\end{table}

\begin{figure*}[htb]
    \centering
    \includegraphics[scale=0.72]{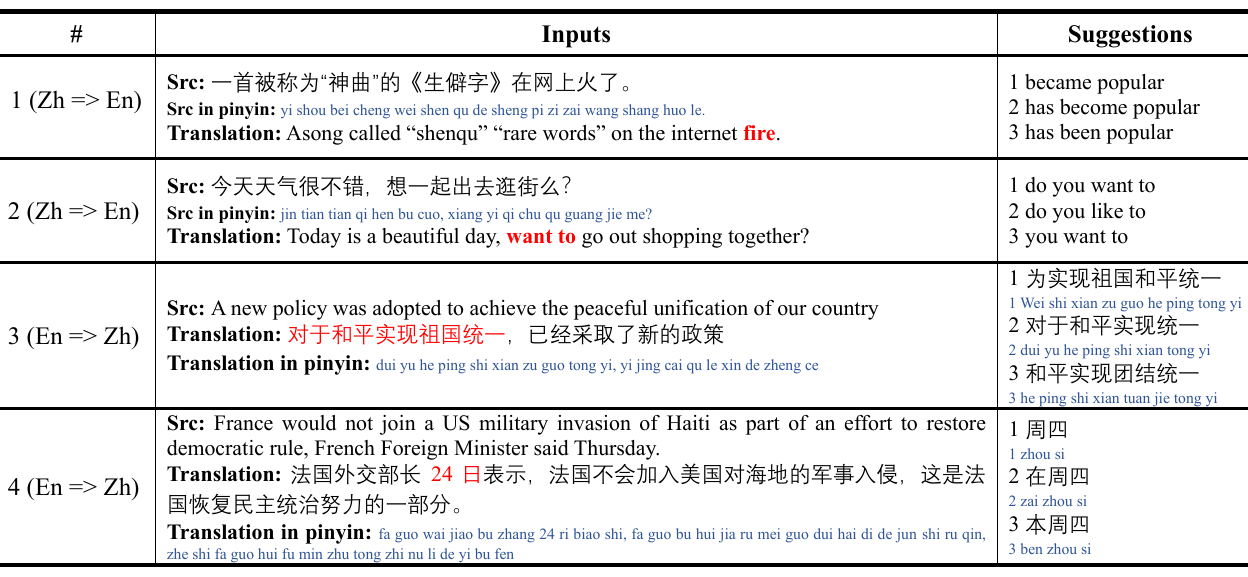}
    \caption {Case study for the proposed approach. 'Src' means the source sentence. The segment in red color represents the incorrect part in the translation, and the top-3 suggestions are provided for each incorrect part. For readability, we provide the pinyin version for each Chinese sentence in blue color.}
    \label{fig:case_study}
\end{figure*}

\vspace{-0.5cm}
\subsection{Case Study}
\label{sec:case_study}
We perform case study in Chinese-English translation directions, and each case includes the source sentence, translation, incorrect word span, and suggestions. For case 1 in Figure \ref{fig:case_study}, the Chinese word ``火了" (means getting popular) has been wrongly translated into its superficial meaning ``fire", and the proposed model gives the right suggestions when the translator selects ``fire" as the incorrect part. Similarly, in case 4, the English word ``Thursday" has been wrongly translated into ``24日", and our model provides three correct alternatives. As for case 2, some important constituents are missed in the translation, which makes the translation not in accordance with the rules of grammar. By selecting the words (``want to") neighboring to the position where there are missing constituents, our model can fill in the missed constituents rightly.
Case 3 demonstrates that the proposed model can generate more fluent alternatives. 

\section{Related Work}
\paragraph{Related tasks.}
Some similar techniques have been explored in CAT. \citet{green2014human} and \citet{knowles2016neural} study the task of so-called translation prediction, which provides predictions of the next word (or phrase) given a prefix. \citet{huang2015new} and \citet{santy2019inmt} further consider the hints of the translator in the task of translation prediction. Compared to TS, the most significant difference is the strict assumption of the translation context, i.e., the prefix context, which severely impedes the use of their methods under the scenarios of PE. Lexically constrained decoding which completes a translation based on some unordered words, relaxes the constraints provided by human translators from prefixes to general forms \cite{hokamp2017lexically,post2018fast,kajiwara2019negative,susanto2020lexically}. Although it does not need to re-train the model, its low efficiency makes it only applicable in scenarios where only a few constraints need to be applied. Recently, \citet{li2021gwlan} study the problem of auto-completion with different context types. However, they only focus on the word-level auto-completion, and their experiments are also conducted on the automatically constructed datasets.
\paragraph{Related models.}
\citet{lee2021intellicat} propose to perform translation suggestion based on XLM-R, where the model is trained to predict the masked span of the translation sentence. During inference, they need to generate multiple inputs for the selected sequence of words, with each input containing a different number of the "[MASK]" token. Therefore, the inference process of XLM-R based model gets complex and time-consuming. With the success on many sequence-to-sequence tasks, Transformer can generate sequences with various lengths. The naive Transformer treats each token in the input sentence without any distinction. Based on Transformer, \cite{junczys2018ms} propose the dual-source encoder for the task of PE. \citet{wang2020touch} also apply the dual-source encoder to the touch-editing scenario, and they also consider the translator's actions for PE. In parallel to our work, \citet{zhang2021domain} propose a domain-aware self-attention to address the domain adaptation. While their idea is similar to the proposed segment-aware self-attention, they introduce large-scale additional parameters.

\section{Conclusion and Future work}
In this paper, we propose a benchmark for the task of translation suggestion. We construct and share a golden dataset, named \emph{WeTS}, for the community, and propose several ways for automatically constructing the synthetic corpora which can be used to improve the performance substantially. Additionally, we for the first time propose the segment-aware self-attention based Transformer, named SA-Transformer, which achieves the best performance on all four translation directions. We hope our work will provide a new perspective and spur future researches on TS. There are two promising directions for the future work. First, we decide to introduce new techniques from recommendation systems to generate more diverse and accurate suggestions. Second, modeling the interactions  between the source and translation sentences implicitly may be helpful to improve the performance.

\section*{Limitations}
\label{sec:limits}
While achieving promising performance, the proposed model still has some weaknesses in the real application: 1) The suggestions sometimes have low diversity. This is mainly because that the search space of the beam search is too narrow to extract diverse suggestions \cite{wu2020generating,sun2020generating}. This problem could be solved by utilizing some random search strategies during inference or the diverse beam search algorithms \cite{vijayakumar2016diverse}.  
2) The model tends to provide less satisfactory suggestions for the long word spans. Poorer performance for longer sequence is the general weakness for most of the neural models. However, this problem may be more urgent for the proposed TS model since the inputs are much longer than the naive translation model (the concatenation between the source and translation sentences). We believe this problem can be alleviated by using the separate encoders for the source and translation sentences. However, our experimental results show that separate encoders in dual-source Transformer achieves inferior performance as no cross-lingual interactions are modeled. Therefore, a promising direction is that the model encodes the source and translation sentences separately in the lower layers and modeling the interactions in the upper layers.
3) The best suggestion does not always rank in the first position. While the results of the beam search are ranked according to the predicted scores, the ranked positions are not always satisfactory. More features extracted from the source and translation sentences should be considered into the re-ranking process.  

Another potential limitation proposed by the anonymous reviewers is that the in-house data used to train the En-Zh model. There are two reasons that we add the in-house En-Zh data: 1) The in-house En-Zh data can be used to train the model as strong as possible, which makes the benchmark solid for the production; 2) While we did not open the in-house corpus, we have released the trained En-Zh NMT model, which makes it still easy to re-produce the results.

\section*{Acknowledgements}
The authors would like to thank the anonymous reviewers of this paper, and the anonymous reviewers of the previous version for their valuable comments and suggestions to improve our work.

\bibliography{anthology,custom}

\begin{thebibliography}{32}
\expandafter\ifx\csname natexlab\endcsname\relax\def\natexlab#1{#1}\fi

\bibitem[{Alabau et~al.(2014)Alabau, Buck, Carl, Casacuberta,
  Garc{\'\i}a-Mart{\'\i}nez, Germann, Gonz{\'a}lez-Rubio, Hill, Koehn, Leiva
  et~al.}]{alabau2014casmacat}
Vicent Alabau, Christian Buck, Michael Carl, Francisco Casacuberta, Mercedes
  Garc{\'\i}a-Mart{\'\i}nez, Ulrich Germann, Jes{\'u}s Gonz{\'a}lez-Rubio,
  Robin Hill, Philipp Koehn, Luis~A Leiva, et~al. 2014.
\newblock Casmacat: A computer-assisted translation workbench.
\newblock In \emph{Proceedings of the Demonstrations at the 14th Conference of
  the European Chapter of the Association for Computational Linguistics}, pages
  25--28.

\bibitem[{Bahdanau et~al.(2015)Bahdanau, Cho, and Bengio}]{bahdanau:14}
Dzmitry Bahdanau, Kyunghyun Cho, and Yoshua Bengio. 2015.
\newblock Neural machine translation by jointly learning to align and
  translate.
\newblock In \emph{3rd International Conference on Learning Representations,
  {ICLR} 2015, San Diego, CA, USA, May 7-9, 2015, Conference Track
  Proceedings}.

\bibitem[{Barrachina et~al.(2009)Barrachina, Bender, Casacuberta, Civera,
  Cubel, Khadivi, Lagarda, Ney, Tom{\'a}s, Vidal
  et~al.}]{barrachina2009statistical}
Sergio Barrachina, Oliver Bender, Francisco Casacuberta, Jorge Civera, Elsa
  Cubel, Shahram Khadivi, Antonio Lagarda, Hermann Ney, Jes{\'u}s Tom{\'a}s,
  Enrique Vidal, et~al. 2009.
\newblock Statistical approaches to computer-assisted translation.
\newblock \emph{Computational Linguistics}, 35(1):3--28.

\bibitem[{Cho et~al.(2014)Cho, van Merrienboer, G{\"u}l{\c{c}}ehre, Bahdanau,
  Bougares, Schwenk, and Bengio}]{cho2014learning}
Kyunghyun Cho, Bart van Merrienboer, {\c{C}}aglar G{\"u}l{\c{c}}ehre, Dzmitry
  Bahdanau, Fethi Bougares, Holger Schwenk, and Yoshua Bengio. 2014.
\newblock Learning phrase representations using rnn encoder-decoder for
  statistical machine translation.
\newblock In \emph{EMNLP}.

\bibitem[{Conneau et~al.(2020)Conneau, Khandelwal, Goyal, Chaudhary, Wenzek,
  Guzm{\'a}n, Grave, Ott, Zettlemoyer, and
  Stoyanov}]{conneau-etal-2020-unsupervised}
Alexis Conneau, Kartikay Khandelwal, Naman Goyal, Vishrav Chaudhary, Guillaume
  Wenzek, Francisco Guzm{\'a}n, Edouard Grave, Myle Ott, Luke Zettlemoyer, and
  Veselin Stoyanov. 2020.
\newblock \href {https://doi.org/10.18653/v1/2020.acl-main.747} {Unsupervised
  cross-lingual representation learning at scale}.
\newblock In \emph{Proceedings of the 58th Annual Meeting of the Association
  for Computational Linguistics}, pages 8440--8451, Online. Association for
  Computational Linguistics.

\bibitem[{CONNEAU and Lample(2019)}]{lample2019cross}
Alexis CONNEAU and Guillaume Lample. 2019.
\newblock \href
  {https://proceedings.neurips.cc/paper/2019/file/c04c19c2c2474dbf5f7ac4372c5b9af1-Paper.pdf}
  {Cross-lingual language model pretraining}.
\newblock In \emph{Advances in Neural Information Processing Systems},
  volume~32. Curran Associates, Inc.

\bibitem[{Dyer et~al.(2013)Dyer, Chahuneau, and Smith}]{dyer2013simple}
Chris Dyer, Victor Chahuneau, and Noah~A Smith. 2013.
\newblock A simple, fast, and effective reparameterization of ibm model 2.
\newblock In \emph{Proceedings of the 2013 Conference of the North American
  Chapter of the Association for Computational Linguistics: Human Language
  Technologies}, pages 644--648.

\bibitem[{Green et~al.(2013)Green, Heer, and Manning}]{green2013efficacy}
Spence Green, Jeffrey Heer, and Christopher~D Manning. 2013.
\newblock The efficacy of human post-editing for language translation.
\newblock In \emph{Proceedings of the SIGCHI conference on human factors in
  computing systems}, pages 439--448.

\bibitem[{Green et~al.(2014)Green, Wang, Chuang, Heer, Schuster, and
  Manning}]{green2014human}
Spence Green, Sida~I Wang, Jason Chuang, Jeffrey Heer, Sebastian Schuster, and
  Christopher~D Manning. 2014.
\newblock Human effort and machine learnability in computer aided translation.
\newblock In \emph{Proceedings of the 2014 Conference on Empirical Methods in
  Natural Language Processing (EMNLP)}, pages 1225--1236.

\bibitem[{Hokamp and Liu(2017)}]{hokamp2017lexically}
Chris Hokamp and Qun Liu. 2017.
\newblock \href {https://doi.org/10.18653/v1/P17-1141} {Lexically constrained
  decoding for sequence generation using grid beam search}.
\newblock In \emph{Proceedings of the 55th Annual Meeting of the Association
  for Computational Linguistics (Volume 1: Long Papers)}, pages 1535--1546,
  Vancouver, Canada. Association for Computational Linguistics.

\bibitem[{Huang et~al.(2015)Huang, Zhang, Zhou, and Zong}]{huang2015new}
Guoping Huang, Jiajun Zhang, Yu~Zhou, and Chengqing Zong. 2015.
\newblock A new input method for human translators: integrating machine
  translation effectively and imperceptibly.
\newblock In \emph{Twenty-Fourth International Joint Conference on Artificial
  Intelligence}.

\bibitem[{Jalili~Sabet et~al.(2020)Jalili~Sabet, Dufter, Yvon, and
  Sch{\"u}tze}]{sabet2020simalign}
Masoud Jalili~Sabet, Philipp Dufter, Fran{\c{c}}ois Yvon, and Hinrich
  Sch{\"u}tze. 2020.
\newblock \href {https://doi.org/10.18653/v1/2020.findings-emnlp.147}
  {{S}im{A}lign: High quality word alignments without parallel training data
  using static and contextualized embeddings}.
\newblock In \emph{Findings of the Association for Computational Linguistics:
  EMNLP 2020}, pages 1627--1643, Online. Association for Computational
  Linguistics.

\bibitem[{Junczys-Dowmunt and Grundkiewicz(2018)}]{junczys2018ms}
Marcin Junczys-Dowmunt and Roman Grundkiewicz. 2018.
\newblock \href {https://doi.org/10.18653/v1/W18-6467} {{MS}-{UE}din submission
  to the {WMT}2018 {APE} shared task: Dual-source transformer for automatic
  post-editing}.
\newblock In \emph{Proceedings of the Third Conference on Machine Translation:
  Shared Task Papers}, pages 822--826, Belgium, Brussels. Association for
  Computational Linguistics.

\bibitem[{Kajiwara(2019)}]{kajiwara2019negative}
Tomoyuki Kajiwara. 2019.
\newblock Negative lexically constrained decoding for paraphrase generation.
\newblock In \emph{Proceedings of the 57th Annual Meeting of the Association
  for Computational Linguistics}, pages 6047--6052.

\bibitem[{Knowles and Koehn(2016)}]{knowles2016neural}
Rebecca Knowles and Philipp Koehn. 2016.
\newblock Neural interactive translation prediction.
\newblock In \emph{Proceedings of the Association for Machine Translation in
  the Americas}, pages 107--120.

\bibitem[{Lai et~al.(2022)Lai, Yang, Meng, Chen, Xu, and Zhou}]{Lai2022cross}
Siyu Lai, Zhen Yang, Fandong Meng, Yufeng Chen, Jinan Xu, and Jie Zhou. 2022.
\newblock \href {https://arxiv.org/abs/2210.04141} {Cross-align: Modeling deep
  cross-lingual interactions for word alignment}.
\newblock \emph{arXiv preprint arXiv:2210.04141}.

\bibitem[{Lee et~al.(2021)Lee, Ahn, Park, and Jo}]{lee2021intellicat}
Dongjun Lee, Junhyeong Ahn, Heesoo Park, and Jaemin Jo. 2021.
\newblock \href {https://doi.org/10.18653/v1/2021.acl-demo.2} {{I}ntelli{CAT}:
  Intelligent machine translation post-editing with quality estimation and
  translation suggestion}.
\newblock In \emph{Proceedings of the 59th Annual Meeting of the Association
  for Computational Linguistics and the 11th International Joint Conference on
  Natural Language Processing: System Demonstrations}, pages 11--19, Online.
  Association for Computational Linguistics.

\bibitem[{Li et~al.(2021)Li, Liu, Huang, and Shi}]{li2021gwlan}
Huayang Li, Lemao Liu, Guoping Huang, and Shuming Shi. 2021.
\newblock \href {https://doi.org/10.18653/v1/2021.acl-long.370} {{GWLAN}:
  General word-level {A}utocompletio{N} for computer-aided translation}.
\newblock In \emph{Proceedings of the 59th Annual Meeting of the Association
  for Computational Linguistics and the 11th International Joint Conference on
  Natural Language Processing (Volume 1: Long Papers)}, pages 4792--4802,
  Online. Association for Computational Linguistics.

\bibitem[{Papineni et~al.(2002)Papineni, Roukos, Ward, and
  Zhu}]{papineni2002bleu}
Kishore Papineni, Salim Roukos, Todd Ward, and Wei-Jing Zhu. 2002.
\newblock Bleu: a method for automatic evaluation of machine translation.
\newblock In \emph{Proceedings of the 40th annual meeting of the Association
  for Computational Linguistics}, pages 311--318.

\bibitem[{Post and Vilar(2018)}]{post2018fast}
Matt Post and David Vilar. 2018.
\newblock \href {https://doi.org/10.18653/v1/N18-1119} {Fast lexically
  constrained decoding with dynamic beam allocation for neural machine
  translation}.
\newblock In \emph{Proceedings of the 2018 Conference of the North {A}merican
  Chapter of the Association for Computational Linguistics: Human Language
  Technologies, Volume 1 (Long Papers)}, pages 1314--1324, New Orleans,
  Louisiana. Association for Computational Linguistics.

\bibitem[{Santy et~al.(2019)Santy, Dandapat, Choudhury, and
  Bali}]{santy2019inmt}
Sebastin Santy, Sandipan Dandapat, Monojit Choudhury, and Kalika Bali. 2019.
\newblock Inmt: Interactive neural machine translation prediction.
\newblock In \emph{Proceedings of the 2019 Conference on Empirical Methods in
  Natural Language Processing and the 9th International Joint Conference on
  Natural Language Processing (EMNLP-IJCNLP): System Demonstrations}, pages
  103--108.

\bibitem[{Sellam et~al.(2020)Sellam, Das, and Parikh}]{sellam2020bleurt}
Thibault Sellam, Dipanjan Das, and Ankur Parikh. 2020.
\newblock \href {https://doi.org/10.18653/v1/2020.acl-main.704} {{BLEURT}:
  Learning robust metrics for text generation}.
\newblock In \emph{Proceedings of the 58th Annual Meeting of the Association
  for Computational Linguistics}, pages 7881--7892, Online. Association for
  Computational Linguistics.

\bibitem[{Sennrich et~al.(2016)Sennrich, Haddow, and
  Birch}]{sennrich-etal-2016-neural}
Rico Sennrich, Barry Haddow, and Alexandra Birch. 2016.
\newblock Neural machine translation of rare words with subword units.
\newblock In \emph{Proceedings of the 54th Annual Meeting of the Association
  for Computational Linguistics (Volume 1: Long Papers)}, pages 1715--1725,
  Berlin, Germany. Association for Computational Linguistics.

\bibitem[{Snover et~al.(2006)Snover, Dorr, Schwartz, Micciulla, and
  Makhoul}]{snover2006study}
Matthew Snover, Bonnie Dorr, Richard Schwartz, Linnea Micciulla, and John
  Makhoul. 2006.
\newblock A study of translation edit rate with targeted human annotation.
\newblock In \emph{Proceedings of the 7th Conference of the Association for
  Machine Translation in the Americas: Technical Papers}, pages 223--231.

\bibitem[{Sun et~al.(2020)Sun, Huang, Wei, Dai, and Chen}]{sun2020generating}
Zewei Sun, Shujian Huang, Hao-Ran Wei, Xin-yu Dai, and Jiajun Chen. 2020.
\newblock Generating diverse translation by manipulating multi-head attention.
\newblock In \emph{Proceedings of the AAAI Conference on Artificial
  Intelligence}, volume~34, pages 8976--8983.

\bibitem[{Susanto et~al.(2020)Susanto, Chollampatt, and
  Tan}]{susanto2020lexically}
Raymond~Hendy Susanto, Shamil Chollampatt, and Liling Tan. 2020.
\newblock \href {https://doi.org/10.18653/v1/2020.acl-main.325} {Lexically
  constrained neural machine translation with {L}evenshtein transformer}.
\newblock In \emph{Proceedings of the 58th Annual Meeting of the Association
  for Computational Linguistics}, pages 3536--3543, Online. Association for
  Computational Linguistics.

\bibitem[{Vaswani et~al.(2017)Vaswani, Shazeer, Parmar, Uszkoreit, Jones,
  Gomez, Kaiser, and Polosukhin}]{vaswani2017attention}
Ashish Vaswani, Noam Shazeer, Niki Parmar, Jakob Uszkoreit, Llion Jones,
  Aidan~N Gomez, {\L}ukasz Kaiser, and Illia Polosukhin. 2017.
\newblock Attention is all you need.
\newblock In \emph{Advances in neural information processing systems}, pages
  5998--6008.

\bibitem[{Vijayakumar et~al.(2016)Vijayakumar, Cogswell, Selvaraju, Sun, Lee,
  Crandall, and Batra}]{vijayakumar2016diverse}
Ashwin~K Vijayakumar, Michael Cogswell, Ramprasath~R Selvaraju, Qing Sun,
  Stefan Lee, David Crandall, and Dhruv Batra. 2016.
\newblock Diverse beam search: Decoding diverse solutions from neural sequence
  models.
\newblock \emph{arXiv preprint arXiv:1610.02424}.

\bibitem[{Wang et~al.(2020)Wang, Zhang, Liu, Huang, and Zong}]{wang2020touch}
Qian Wang, Jiajun Zhang, Lemao Liu, Guoping Huang, and Chengqing Zong. 2020.
\newblock Touch editing: A flexible one-time interaction approach for
  translation.
\newblock In \emph{Proceedings of the 1st Conference of the Asia-Pacific
  Chapter of the Association for Computational Linguistics and the 10th
  International Joint Conference on Natural Language Processing}, pages 1--11.

\bibitem[{Wu et~al.(2020)Wu, Feng, and Shao}]{wu2020generating}
Xuanfu Wu, Yang Feng, and Chenze Shao. 2020.
\newblock \href {https://doi.org/10.18653/v1/2020.emnlp-main.82} {Generating
  diverse translation from model distribution with dropout}.
\newblock In \emph{Proceedings of the 2020 Conference on Empirical Methods in
  Natural Language Processing (EMNLP)}, pages 1088--1097, Online. Association
  for Computational Linguistics.

\bibitem[{Zhang et~al.(2021)Zhang, Liu, Xiong, Zhang, and
  Chen}]{zhang2021domain}
Shiqi Zhang, Yan Liu, Deyi Xiong, Pei Zhang, and Boxing Chen. 2021.
\newblock Domain-aware self-attention for multi-domain neural machine
  translation.
\newblock \emph{Proc. Interspeech 2021}, pages 2047--2051.

\bibitem[{Zouhar et~al.(2021)Zouhar, Popel, Bojar, and
  Tamchyna}]{zouhar2021neural}
Vil{\'e}m Zouhar, Martin Popel, Ond{\v{r}}ej Bojar, and Ale{\v{s}} Tamchyna.
  2021.
\newblock \href {https://doi.org/10.18653/v1/2021.emnlp-main.801} {Neural
  machine translation quality and post-editing performance}.
\newblock In \emph{Proceedings of the 2021 Conference on Empirical Methods in
  Natural Language Processing}, pages 10204--10214, Online and Punta Cana,
  Dominican Republic. Association for Computational Linguistics.

\end{thebibliography}
\bibliographystyle{acl_natbib}

\appendix
\begin{figure*}[htb]
    \centering
    \includegraphics[scale=0.60]{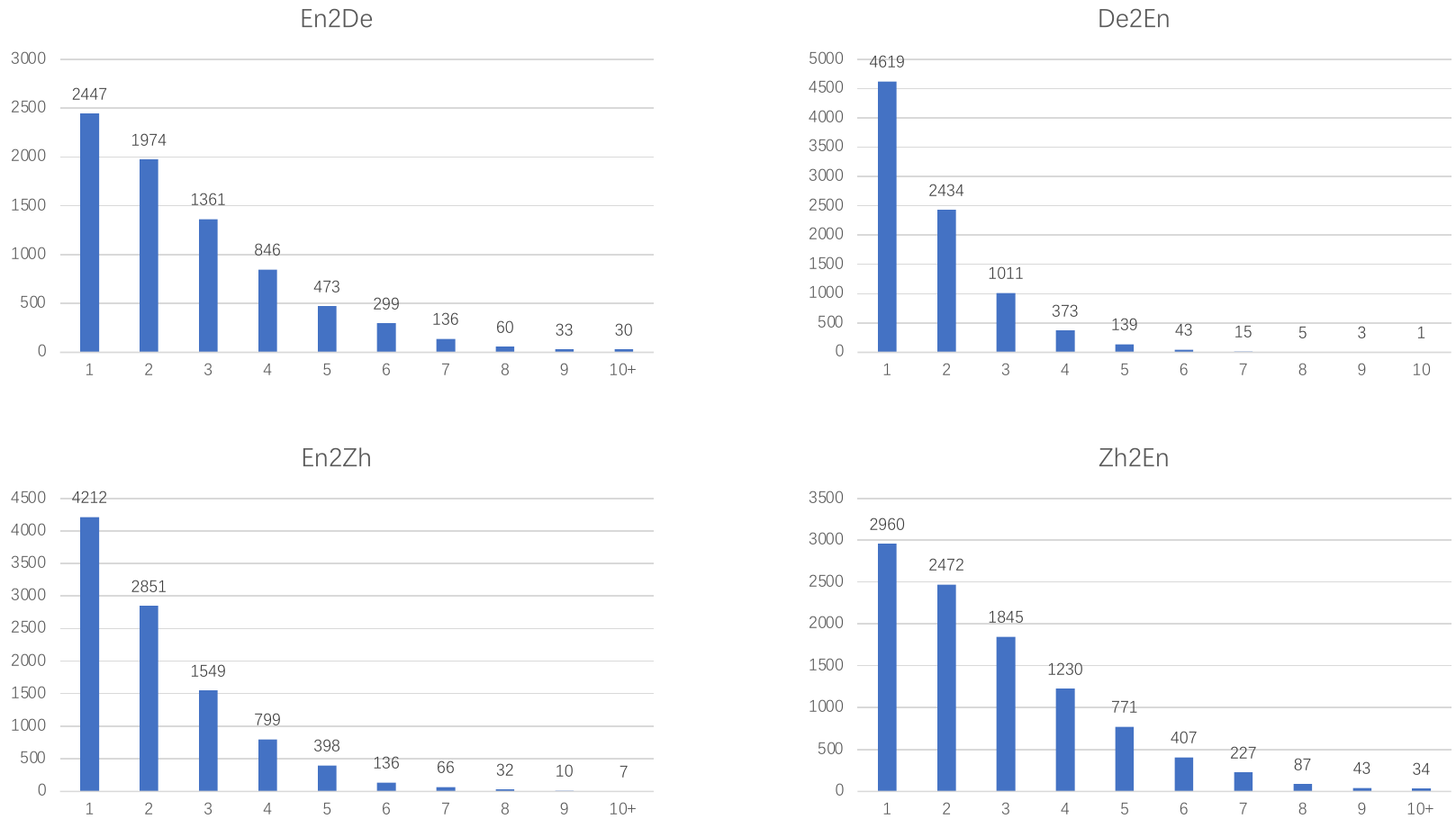}
    \caption {The number of incorrect span in each annotated example. The horizontal axis represents the number of incorrect span in each example, and the the vertical axis denotes the number of annotated examples.}
    \label{fig:n_span}
\end{figure*}

\begin{figure*}[htb]
    \centering
    \includegraphics[scale=0.60]{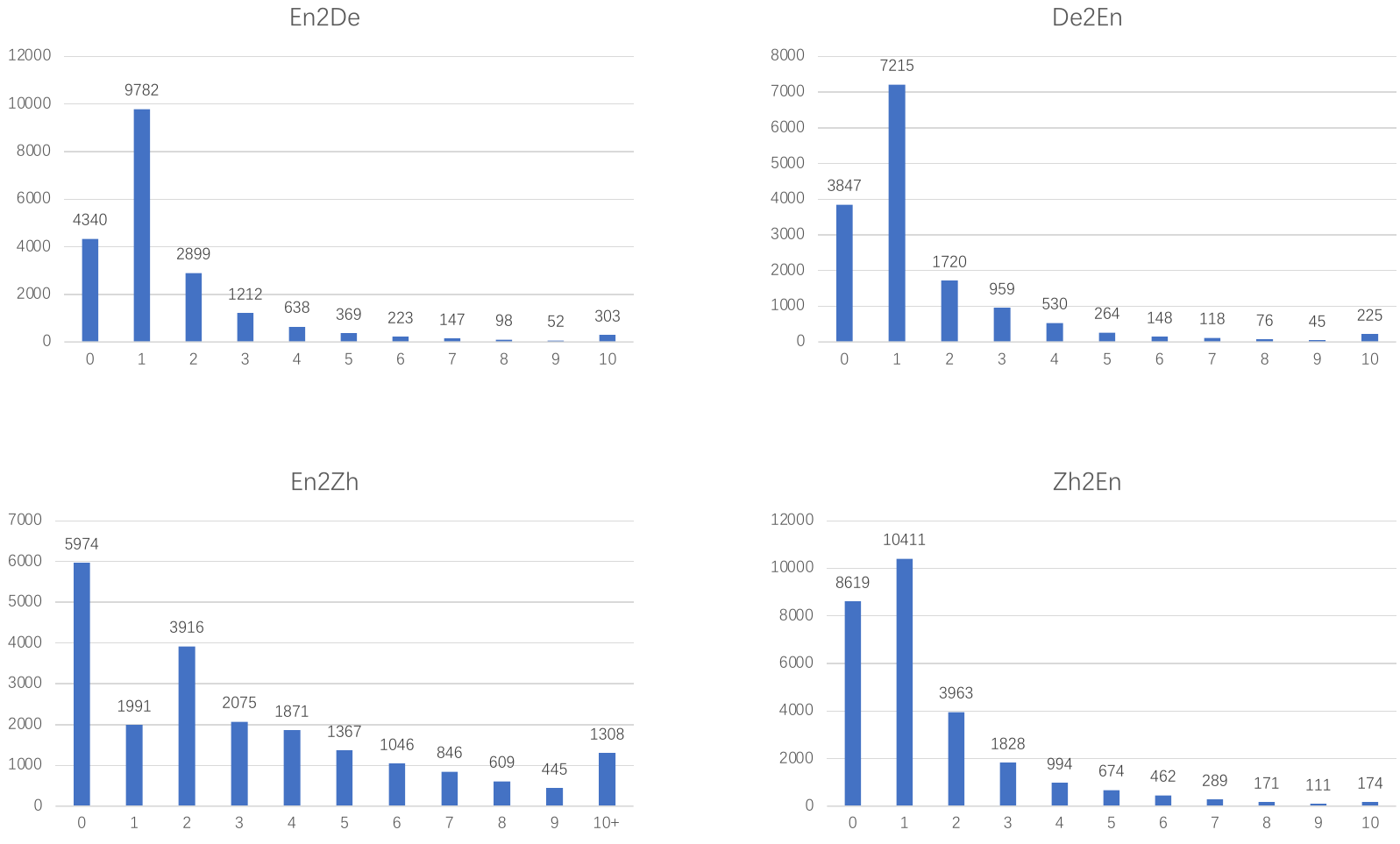}
    \caption {The length of the incorrect span. The horizontal axis represents the length of the incorrect span, and ``0'' means that no incorrect word should be selected but some words should be inserted for the under-translation. The vertical axis denotes the number of the incorrect spans. For Chinese, the length is calculated as the number of Chinese characters included in the incorrect span. For other languages, length is calculated as the number of words.}
    \label{fig:l_span}
\end{figure*}

\begin{figure*}[htb]
    \centering
    \includegraphics[scale=0.60]{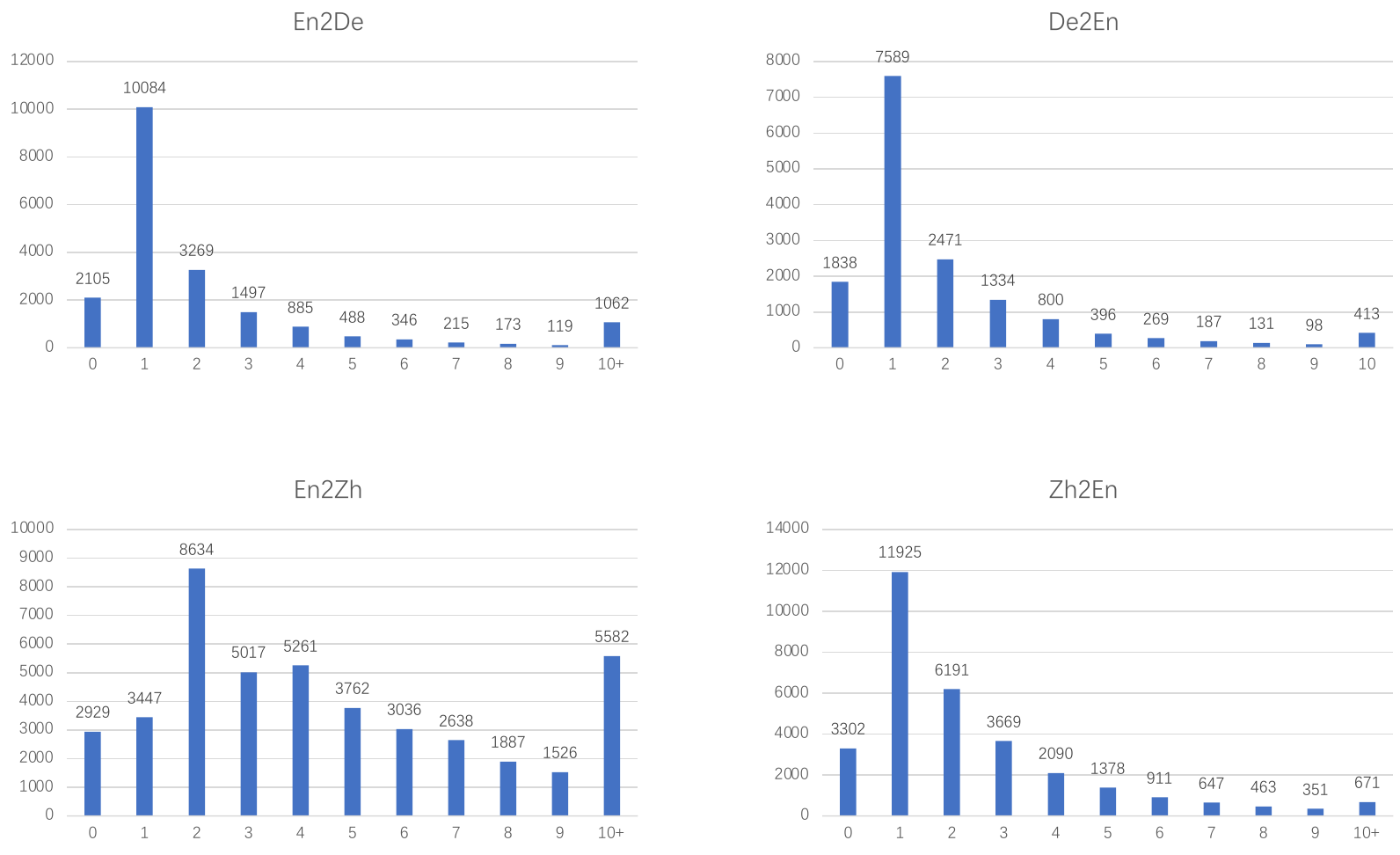}
    \caption {The length of the suggestion.  The horizontal axis represents the length of the suggestion, and ``0'' means that the corresponding the incorrect span should be deleted as the over-translation happens. The vertical axis denotes the number of the suggestions. For Chinese, the length is calculated as the number of Chinese characters included in the incorrect span. For other languages, length is calculated as the number of words. }
    \label{fig:l_sugg}
\end{figure*}

\section{Detailed statistics about \emph{WeTS}}
\label{app:statistics}
This section presents the detailed statistics about the proposed \emph{WeTS}. Since the training, validation and test sets are three homogeneous splits from the randomly shuffled annotated corpus, we only report the statistics on the training set.

\subsection{The number of the incorrect span}
Each annotated example may contain multiple incorrect spans, we show the number of the incorrect span in each annotated example as Figure \ref{fig:n_span}. We can see that most examples have only a few incorrect spans, and there are more than 70 percent examples containing less than 3 incorrect spans for each translation direction.  

\subsection{The length of the incorrect span}
Figure \ref{fig:l_span} represents the length distribution of the incorrect spans. We can find that most of the incorrect spans contain less than 3 words or Chinese characters. This is mainly because of our key rule for annotating the incorrect span as local as possible. Additionally, for all of the four translation directions, the number of the incorrect spans with length 0 ranks top-2 among all the length buckets. This shows that under-translation is still a frequent error of the existing NMT models. 

\subsection{The length of the suggestions}
Figure \ref{fig:l_sugg} shows the length distribution of the suggestions. We can see that in English-to-German, German-to-English and Chinese-to-English, most of the suggestions contain only one word. For English-to-Chinese, most suggestions contain two Chinese characters. Additionally, we can also find that there are quite a few of suggestions with length zero in each translation direction. This shows that over-translation is a non-negligible problem for the existing NMT models.

\section{Pre-processing in detail}
\label{app:processing}
For learning the BPE codes on Chinese-English language pairs, the number of the merge operation is set as 64,000. For English-German language pairs, the number of merge operation is 32,000. For constructing the synthetic corpora, we perform randomly sampling on the golden and pseudo parallel corpus. The sizes of the constructed synthetic corpora are listed as Table \ref{tab:statics of syn}.
\begin{table}[ht]
			\centering
			\resizebox{0.95 \columnwidth}{!}{
				\begin{tabular}{c|ccc}
					\toprule[2pt]
					 Directions & on golden & on pseudo & with word alignment \\
					\midrule[1pt]
			        En$\Rightarrow$De & 9.0M & 9.0M & 5.8M \\
			        De$\Rightarrow$En & 9.0M & 9.0M & 5.3M \\
			        Zh$\Rightarrow$En & 20M & 20M & 19.2M  \\
			        En$\Rightarrow$Zh & 20M & 20M & 18.4M  \\
					\bottomrule[2pt]
				\end{tabular}}
					\caption{\label{tab:statics of syn} The sizes about the constructed synthetic corpora. "on golden" indicates the method of sampling on the golden parallel corpus.}
\end{table}

\section{Experimental settings in detail}
\label{app:settings}
Following the base model in \citet{vaswani2017attention}, we set the word embedding as 512, dropout rate as 0.1 and the head number as 8. We use beam search with a beam size of 4. The proposed model is implemented based on the open-source toolkit fairseq.\footnote{\url{https://github.com/pytorch/fairseq}} For generating the synthetic corpus with word alignment, we set $\beta$ as 10. During pre-training, the batch size is set as 81,920 tokens, and the learning rate is set as 0.0008. During fine-tuning, the batch size and learning rate are set as 41,960 and 0.0001 respectively. For the first-phase pre-training, we stop training when the model achieves no improvements for the tenth evaluation on the development set. For the process of second-phase pre-training and fine-tuning, we train the whole model for 200,000 and 100 steps respectively. For calculating the BLEURT score, we use the default models (BLEURT-20) provided by the bleurt toolkit.  


 \end{CJK} 
\end{document}